\documentclass{article}
\usepackage{graphicx} 
\usepackage{makecell}
\usepackage{booktabs}
\usepackage{url}
\usepackage[table]{xcolor}
\definecolor{mylight}{rgb}{0.95, 0.95, 1.0}  
\usepackage{cite}
\usepackage{fontawesome}
\usepackage{multirow}
\usepackage{tablefootnote}
\usepackage{arxiv}
\usepackage{hyperref}

\title{ocean-ocr: towards general ocr application \\ via a vision-language model}

\author{
    Song Chen$^{1}$\thanks{Equal core contributors. }\enskip\enskip
    Xinyu Guo$^{1}$\footnotemark[1]\enskip\enskip
    Yadong Li$^{1}$\footnotemark[1]\enskip\enskip 
    Tao Zhang$^{1}$\enskip\enskip
    Mingan Lin$^{1}$\enskip\enskip 
    Dongdong Kuang$^{1,2}$\enskip\enskip \\
    Youwei Zhang$^{1,3}$\enskip\enskip
    Lingfeng Ming$^{1}$\enskip\enskip 
    Fengyu Zhang$^{1}$\enskip\enskip
    Yuran Wang$^{1,4}$\enskip\enskip
    Jianhua Xu$^{1}$\thanks{Corresponding author.}\enskip\enskip
    Zenan Zhou$^{1}\footnotemark[2]$\enskip\enskip
    Weipeng Chen$^{1}$\enskip\enskip \\
    \textsuperscript{1} Baichuan Inc. \enskip
    \textsuperscript{2} Beihang University \enskip
    \textsuperscript{3} Beijing University of Posts and Telecommunications \enskip 
    \textsuperscript{4} Wuhan University \enskip 
    \\
    \texttt{\{xujianhua, zhouzenan\}@baichuan-inc.com}
    \enskip \\
}

\begin{document}

\maketitle

\begin{center}
  \vspace{-3em}
  \faGithub~\url{https://github.com/guoxy25/Ocean-OCR}
  \vspace{0.75em}
\end{center}

\begin{figure}[ht]
\centering
\includegraphics[width=0.92\textwidth]{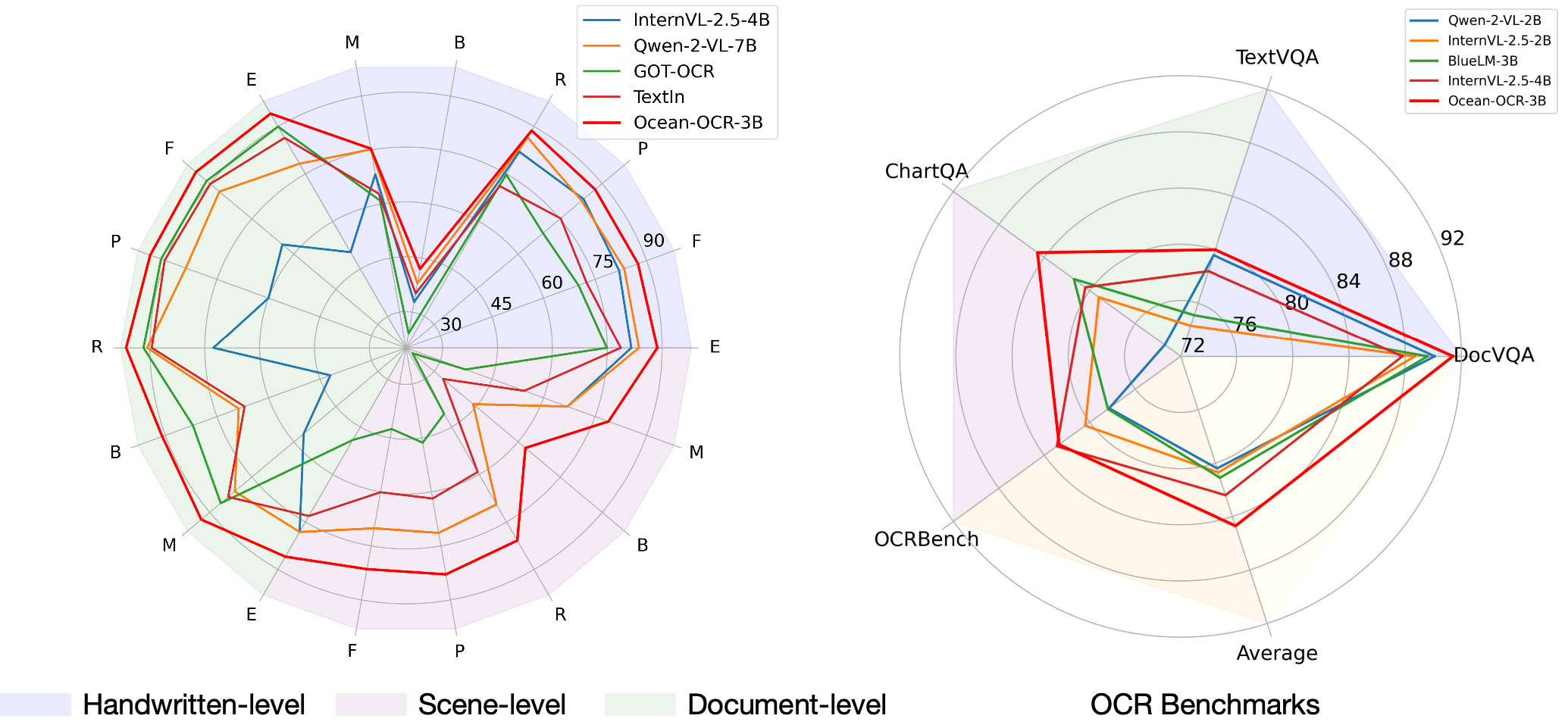}
\vspace{-2pt}
\caption{\textbf{Comparison with models across various OCR scenarios and benchmarks.}
\textbf{(Left)} 
Current mainstream MLLMs and specific OCR models across multiple noteworthy OCR abilities, that is, scene-level, document-level, and handwritten-level text recognition. \textit{E}, \textit{F}, \textit{P}, \textit{R}, \textit{B}, and \textit{M} are the abbreviations for \textit{Edit Distance}, \textit{F1-Score}, \textit{Precision}, \textit{Recall}, \textit{BLEU}, and \textit{METEOR} respectively.
For \textit{Edit Distance}, the plotted score is computed with $x_{after} = 100 - x_{before}$ for better visualization. \textbf{(Right)} Comparison of mainstream MLLMs performance on OCR benchmarks.}
\end{figure}

\begin{abstract}
Multimodal large language models (MLLMs) have shown impressive capabilities across various domains, excelling in processing and understanding information from multiple modalities. Despite the rapid progress made previously, insufficient OCR ability hinders MLLMs from excelling in text-related tasks. In this paper, we present \textbf{Ocean-OCR}, a 3B MLLM with state-of-the-art performance on various OCR scenarios and comparable understanding ability on general tasks. We employ Native Resolution ViT to enable variable resolution input and utilize a substantial collection of high-quality OCR datasets to enhance the model performance.
We demonstrate the superiority of Ocean-OCR through comprehensive experiments on open-source OCR benchmarks and across various OCR scenarios. These scenarios encompass document understanding, scene text recognition, and handwritten recognition, highlighting the robust OCR capabilities of Ocean-OCR.
Note that Ocean-OCR is the first MLLM to outperform professional OCR models such as TextIn and PaddleOCR.

\end{abstract}

\section{Introduction}

Recently, multimodal large language models (MLLMs)\cite{minigemini, survey, internlm, vila, qwen2, internvl2d5} have risen to prominence as a crucial advancement in artificial intelligence, demonstrating the ability to process and understand information across various modalities, including text, images, and videos. 
Nonetheless, developing effective MLLMs remains challenging, which demands sophisticated architectures, comprehensive and high-quality data, and extensive computational resources.
Besides, as a vital source of information, recognition of textual content within images asks for a more fine-grained perception, posing a significant obstacle for improving the OCR ability of MLLMs. 

Various attempts have been made to empower the OCR ability of MLLMs, including the sliding window strategy\cite{monkey, textmonkey}, the layout-aware compression\cite{mplug1d5, mplug2}, etc.
MLLMs mainly focus on visual reasoning performance, and consequently, their capabilities in perception are not as strong. Given this limitation, some studies argue that MLLMs are not well-suited for OCR tasks\cite{got}. However, our Ocean-OCR model delivers exceptional OCR performance while retaining powerful reasoning abilities.

In this work, we introduce Ocean-OCR, a 3B MLLM that excels in OCR tasks while achieving comparable performance on general-purpose tasks. Ocean-OCR adopts the Native Resolution ViT (NaViT)\cite{navit} to address the challenge of varying resolutions present in OCR tasks and employs an MLP to map the visual tokens into the language feature space.
We assess Ocean-OCR across a diverse set of comprehensive benchmarks to highlight its broad applicability and robust general-purpose performance.
On various OCR-related benchmarks, our model consistently exhibits superior performance, showcasing a clear advantage over other models.
To evaluate the OCR ability in real-world applications, we construct extensive evaluation datasets including bilingual dense document understanding, practical scene text recognition, and bilingual handwritten text recognition. In these real-world OCR scenarios, our Ocean-OCR exhibits significantly leading performance.
The key advances in Ocean-OCR include the following:


\begin{itemize}
    \item We introduce Ocean-OCR, a versatile MLLM with 3B parameters that accommodates visual inputs of any resolution. Ocean-OCR is the first MLLM to outperform professional OCR models such as TextIn and PaddleOCR in various OCR scenarios.
    \item Remarkable performance on various benchmarks. Ocean-OCR demonstrates state-of-the-art performance on a multitude of OCR-related benchmarks, such as DocVQA, ChartQA, TextVQA, and OCRBench. Besides, Ocean-OCR also achieves comparable results among mainstream MLLMs with similar parameter sizes on general benchmarks, such as SEEDBench.
    \item Excellence capabilities in real-world OCR applications.
    We construct comprehensive evaluation datasets covering a wide range of OCR application scenarios. Our model outperforms both previous MLLMs and traditional OCR models in all scenarios.
\end{itemize}

\section{Related Work}
\subsection{General MLLMs}

Large Language Models (LLMs) have exhibited remarkable performance across a wide range of downstream tasks. Building on this progress, Multimodal Large Language Models (MLLMs) integrate vision and language information, equipping LLMs with the ability to process multimodal input. To achieve this, researchers have developed methods such as linear projection \cite{llava, qwen2vl}, Q-Former \cite{blip2}, and Perceiver Resampler \cite{flamingo}, each designed to effectively combine visual and textual data. LLaVA series \cite{llava, llava_next, llava1d5} scale the resolution by splitting the image into grids using multiple grid configuration. Intern-VL \cite{internvl, internvl1d5, internvl2d5} series employ dynamic high resolution to capture detailed information and pixel unshuffle strategy to reduce the visual tokens. Qwen2-VL \cite{qwen2vl} introduces the Naive Dynamic Resolution mechanism to dynamically process images of varying resolutions. Despite these efforts, current MLLMs still face challenges in capturing fine-grained information within dense images, particularly in OCR tasks that involve the recognition of complex and densely packed text.

\subsection{MLLM-driven OCR}

Despite the robust perception and reasoning capabilities of current MLLMs, the increasing demand for text-driven visual understanding necessitates more accurate OCR results. UReader introduces a Shape-Adaptive Cropping Module that divides the original image into multiple low-resolution sub-images and employs a shared low-resolution encoder. Monkey \cite{monkey} and TextMonkey \cite{textmonkey} handle high-resolution OCR images by dividing them into patches. mPLUG-DocOwl \cite{mplug, mplug1d5, mplug2} series explore the image cropping and visual token compression approach for document understanding. Vary \cite{vary} introduces an additional SAM-style \cite{sam} visual vocabulary tailored for document and chart data, which operates alongside the CLIP \cite{clip} branch. TextHawk2 \cite{yu2024texthawk2} designs a resampler that can compress visual tokens by a factor of 16 tokens per image. GOT-OCR \cite{got} proposes an innovative paradigm for OCR, which yielded impressive results. Although these methods have offered valuable inspiration to improve the OCR capabilities of MLLMs, their OCR performance still falls short of practical application requirements. In addition, the applicability of these models remains limited to OCR-specific scenarios and does not extend to general-purpose use.

\section{Method}
In this section, we provide a comprehensive overview of Ocean-OCR, delving into the details of its model architecture and foundational components.

\begin{figure}[!t]
\centering
\includegraphics[width=0.8\textwidth]{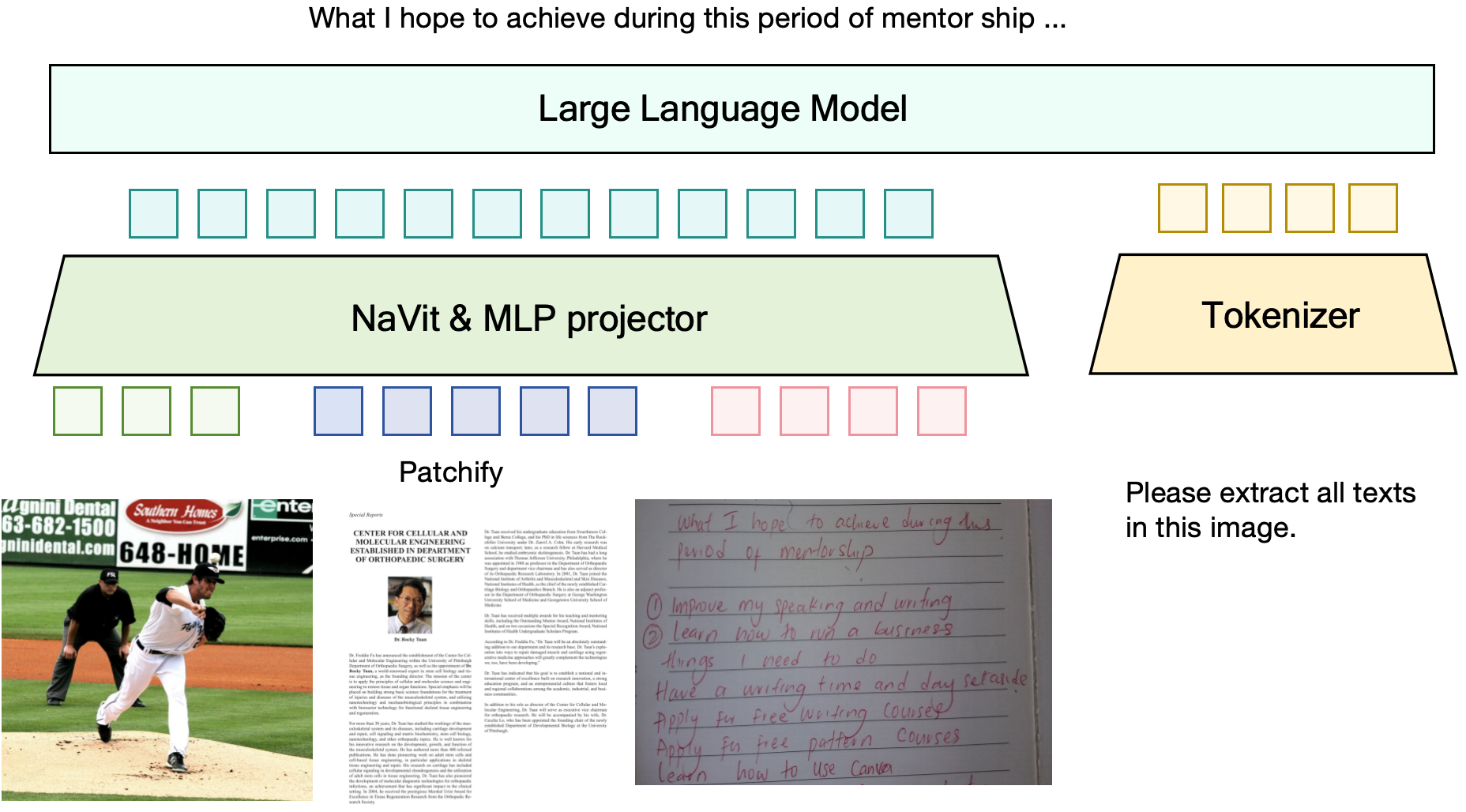}
\caption{Overview of Ocean-OCR-3B. Following most of current MLLMs \cite{qwen2,llava_next,liu2024points1}, Ocean-OCR-3B uses the conventional LLaVA-style structure that consists of a vision encoder, a MLP projector, and a LLM. To better support native dynamic high resolution in various OCR scenarios, we use NaViT-style \cite{navit} vision encoder.}
\label{Fig2}
\end{figure}

\subsection{Basic Framework}

The overall architecture of Ocean-OCR is shown in Fig. \ref{Fig2}, which is composed of the following components. 

\noindent \textbf{Dynamic Resolution and Image Encoder.} Ocean-OCR employs Native Resolution ViT (NaViT) \cite{navit} as visual encoder. 
The visual encoder supports dynamic resolution, enabling Ocean-OCR to process images of any resolution and dynamically convert images into a variable number of visual tokens. 
This design ensures that Ocean-OCR can effectively handle a variety of image sizes while maintaining the integrity and detail of the visual information.

\noindent \textbf{MLP Projector.} We utilize a simple MLP layer as the projector to map the visual tokens to the input space of LLM. To address the issue of an excessive number of visual tokens in high-resolution images, we implement a strategy that compresses adjacent 2 × 2 tokens into a single token. This approach reduces the total number of visual tokens, thereby alleviating computational load while preserving essential visual information. 

\noindent \textbf{LLM.} For the LLM of Ocean-OCR, we employ the Qwen-2.5-3B \cite{qwen2.5}. Considering the ease of use and practical deployment, we opted for this 3B model as it strikes a balance between model capability and size. This choice ensures efficient performance while maintaining robust language understanding and generation abilities, making it well-suited for a wide range of applications.

\subsection{High-Quality Multimodal Data}

\begin{table}[!t]
\centering
\footnotesize
\begin{tabular}{@{}llccc@{}}
\toprule
Phase                                    & Type       & Public Datasets  & Public    & In-House \\ \midrule
\multirow{4}{*}{Alignment\&Pretrain}     & Pure-Text   & -      &   -   &  150.7M    \\ 
                                         & Caption     & \cite{li2024densefusion}\cite{kim2022donut}\cite{DreamLIP}\cite{chen2023internvl}  & 33.2M     &  49.1M   \\ 
                                         & Interleaved & \cite{laurenccon2024obelics}  & 19.1M     &  28.7M   \\ 
                                         & OCR         & \cite{hu2024mplug} & 12.4M       &  7.8M    \\ \midrule
\multirow{2}{*}{Supervised fine-tuning}  & General QA  & \cite{laurençon2024matters} & 3.6M       & -        \\ 
                                         & OCR QA       & \cite{tuo2023anytext} & 3M      &  1.9M    \\ \midrule
                                       Total   & -         & - & 71.3M       &  238.2M    \\ 
                                         \bottomrule
\end{tabular}
\vspace{6pt}
\caption{Detailed statistics of the training data of Ocean-OCR-3B.}
\label{table:data_statistics}
\end{table}

We construct a comprehensive high-quality multimodal dataset from multiple sources to power Ocean-OCR-3B.
As shown in Table \ref{table:data_statistics}, the training data cover a wide range of types, such as pure text data, caption data, interleaved data, and OCR data.
The training process can be divided into three distinct stages: (1) vision-language alignment, (2) vision-vanguage pretraining, and (3) supervised fine-tuning. In the following section, we outline the datasets utilized in each of these stages.


\subsubsection{Vision-language alignment and pretraining data}
\label{sec_pretrain}

\noindent \textbf{Pure text data.} To reserve the strong comprehension abilities of the language model, it is necessary to contain pure text data in the training stage.
For the development of a high-caliber text corpus, we collect data from an extensive variety of sources such as web pages, books, scholarly articles, programming code, and additional resources.
Following the data processing protocols outlined in prior research \cite{dong2024baichuanseed,lu2024datasculpt}, we design a meticulous selection process to enhance both the diversity and the quality of our text corpus.
This emphasis on diversity ensures that our training dataset covers a wide array of subjects and linguistic patterns, making it applicable to a multitude of uses. Additionally, our advanced data processing methods are tailored to remove redundancies and eliminate noise, thus amplifying the dataset's informational richness and overall effectiveness.
%
%
For the vision-language pretraining stage of Ocean-OCR, we maintain a ratio of 50\% pure text data and 50\% vision-language data.

\noindent \textbf{Interleaved image-text data.}
To strengthen the model's capability in handling interleaved image-text data, we utilize the open-source OBELICS \cite{laurenccon2024obelics} as base data. Furthermore, a comprehensive in-house dataset is developed to enrich the model's scope of real-world knowledge. We utilize in-house collected books and papers and parse them to generate interleaved image-text data. These data are highly complete, specialized, and knowledge intensive. The ratio of OBELICS to our in-house data is approximately 4: 6.

\noindent \textbf{Image caption data.}
As an essential part in the training of MLLMs, image caption data directly connect the visual content with textual descriptions. We adopt various open-source caption datasets, including DenseFusion-1M \cite{li2024densefusion}, Synthdog \cite{kim2022donut}, DreamLIP \cite{DreamLIP}, InternVL-SA-1B-Caption \cite{chen2024far,chen2023internvl}. In addition, considering the drop-out of OCR information in these caption data, we synthesized extensive image caption data with OCR hints using PaddleOCR \cite{du2021pp} and GPT-4o \cite{gpt4o}.
The images of the synthetic data come form open-source datasets like Wukong \cite{gu2022wukong} and Laion-2B \cite{schuhmann2022laion}.

\noindent \textbf{OCR data.}
To enhance the model's OCR performance, we utilize both open-source and synthetic OCR datasets. The open-source datasets is composed of DocStruct4M \cite{hu2024mplug}, RenderedText \cite{RenderedText}, 
AnyWord-3M \cite{tuo2023anytext}, TinyChartData \cite{hu2024mplug}.
Our synthetic OCR data contains scene data, PDF document data, and bilingual handwritten text recognition in Chinese and English. For the natural scene data, the Chinese and English images are sampled from the Wukong \cite{gu2022wukong} and Laion-2B \cite{schuhmann2022laion} datasets, respectively.
Specifically, we use the PaddleOCR \cite{du2021pp} tools to generate pseudo-ground truth and then utilize GPT-4o to integrate them into the caption.
We crawl PDF document data from in-house E-book data and use pure-text corpus for rendering handwritten text recognition data.

\subsubsection{Supervised fine-tuning}
\label{sec_sft}
The SFT data of Ocean-OCR is composed of open-source data and in-house synthetic data. The following illustrates the details of the SFT data.

\noindent \textbf{General Visual-Question Answering.}
To enhance the general visual-question answering ability of Ocean-OCR, we utilize the open-source dataset Cauldron \cite{laurençon2024matters}. We perform some data filtering strategy to address certain limitations in the open-source data: (1) Open-source data originates from a wide variety of sources, leading to inconsistent response lengths. (2) The OCR quality is often low. (3) There are instances of hallucinations in the data. We concatenate the image and text data into a dialogue template and then use Qwen2-VL-72B\cite{qwen2} to evaluate the accuracy of the response. This process helps filter out any question-answer pairs that are not sufficiently accurate.

\noindent \textbf{OCR data.} 
Given the limited volume of open-source data, particularly for specific OCR tasks, we have expanded our OCR dataset by synthesizing several types of data: (1) Scene OCR Data: We synthesize scene OCR data using sources such as COCO-Text \cite{veit2016coco}, ICDAR2019 ArT \cite{chng2019icdar2019}, and Incidental Scene Text \cite{yao2015incidental}. For this synthesis, we employ GPT-4o \cite{gpt4o} to generate realistic text visual-question-answers within images. (2) Handwritten OCR Data: To create synthetic handwritten OCR data, we utilize a variety of handwriting styles from different fonts. This data is generated based on corpus content to mimic authentic handwritten text. (3) In-House Document PDF Data: We also include data derived from in-house document PDFs to further enrich the dataset with diverse document layouts and content.
This approach ensures that our OCR dataset is more comprehensive and better suited to handle a wide range of OCR challenges, including scene text recognition, handwritten text recognition, and document understanding.

\subsection{Training Pipelines}
The training process of Ocean-OCR is a three-phase pipeline designed to progressively enhance its multimodal capabilities. (1) First, we focus on training the vision-language projector MLP while keeping the vision encoder and the language model parameters fixed.
%
(2) Second, we engage in comprehensive vision-language pre-training using the datasets described in Section \ref{sec_pretrain}. During this stage, all model parameters are unfrozen and trained concurrently to enhance the model's multimodal understanding. (3) Third, we perform supervised fine-tuning with the datasets outlined in Section \ref{sec_sft}. Similar to the pretraining stage, all components of the model are updated. For all of the three stages, we utilize the next token prediction loss on the text tokens to optimize the model.

\noindent \textbf{Vision-language alignment.} Based on the pre-trained Qwen-2.5-3B language model, our main objective is to project the visual tokens to the feature space of text tokens, thereby enhancing the model's capability to process visual inputs effectively. We introduce NaViT as a dynamic vision encoder to accommodate high-resolution images more flexibly. In this stage, the vision encoder and language model remain frozen, we focus on optimizing the vision-language projector.

\noindent \textbf{Vision-language pretraining.}
Since we have established the vision-language alignment, the following stage is vision-language pre-training dedicated to building extensive joint vision-language knowledge across a variety of tasks. During this phase, all modules are trainable, including the NaViT vision encoder, the MLP projector, and the LLM. After the vision-language pretraining, we can boost the model’s multimodal understanding ability while preserving its robust language capabilities in LLM.

\noindent \textbf{Supervised fine-tuning.}
In the supervised fine-tuning phase, the primary goal is to enhance the instruction following ability of Ocean-OCR and maintain the general ability.



\begin{figure}[!htbp]
\centering
\includegraphics[width=0.92\textwidth]{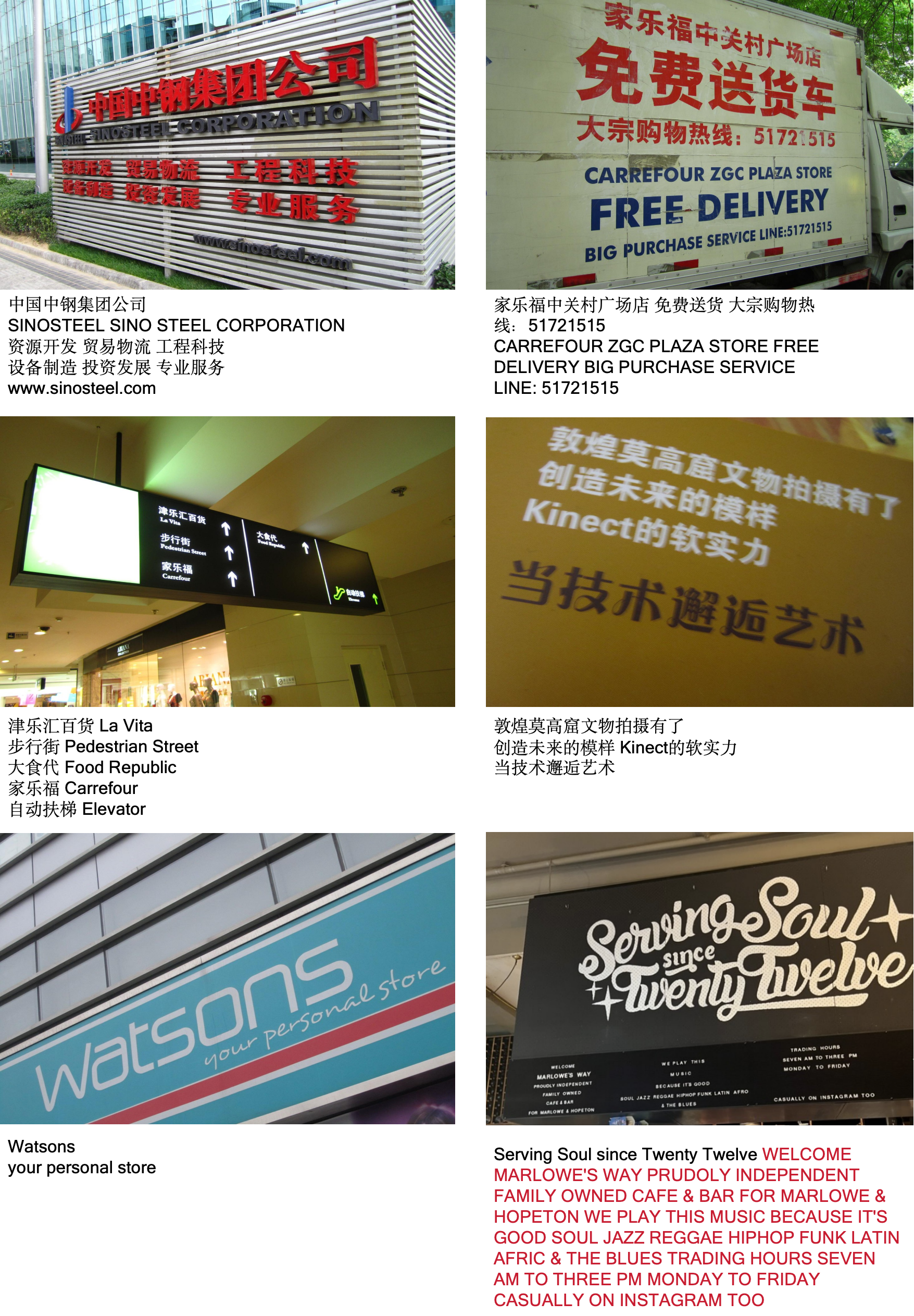}
\caption{Strong OCR ability of Ocean-OCR-3B. Our model shows strong text recognition ability across various real-world scenarios. We simply use \textit{What is written in this image?} as prompt.}
\label{Fig5:scene_ocr}
\end{figure}

\begin{figure}[!htbp]
\centering
\includegraphics[width=0.92\textwidth]{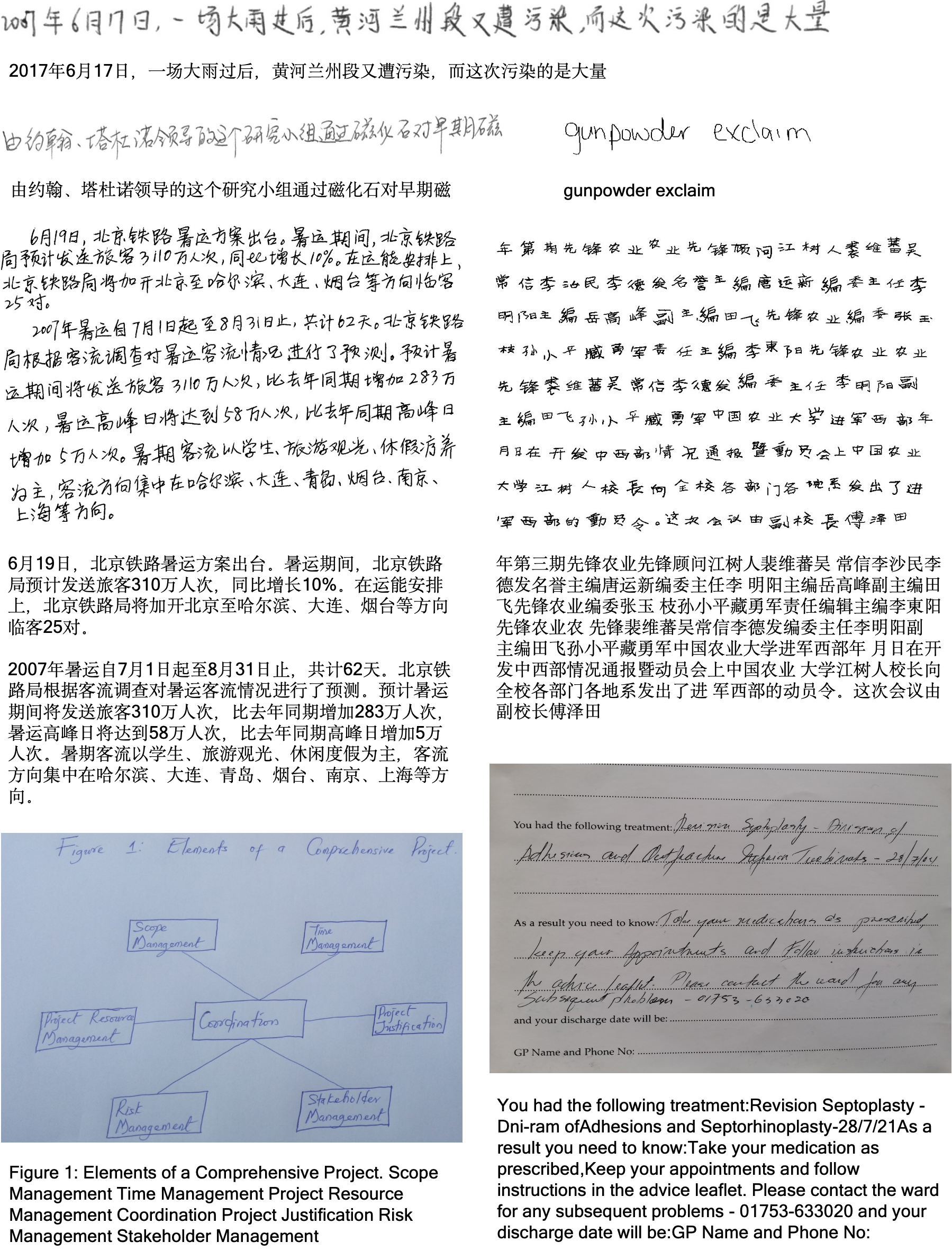}
\caption{Strong OCR ability of Ocean-OCR-3B. Our model shows strong ability for handwritten text recognition in Chinese and English. We simply use \textit{Please extract all texts in this image.} as prompt.}
\label{Fig6:handwritten_ocr}
\end{figure}

\begin{figure}[!htbp]
\centering
\includegraphics[width=0.92\textwidth]{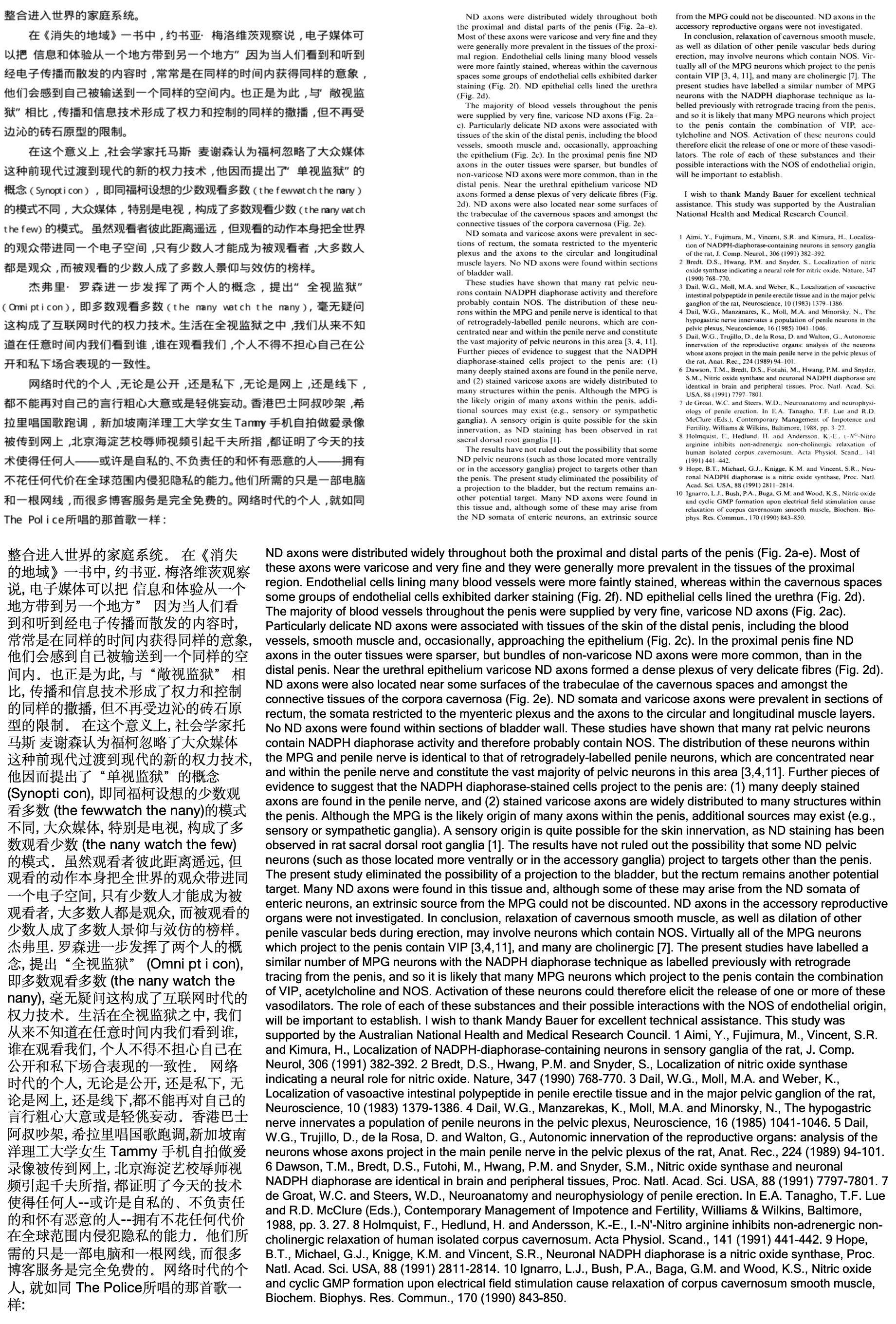}
\caption{Strong OCR ability of Ocean-OCR-3B. Our model shows strong ability for PDF document text recognition in Chinese and English. We simply use \textit{Please extract all texts in this image.} as prompt.}
\label{Fig6:pdf_ocr}
\end{figure}

\begin{figure}[!htbp]
\centering
\includegraphics[width=0.92\textwidth]{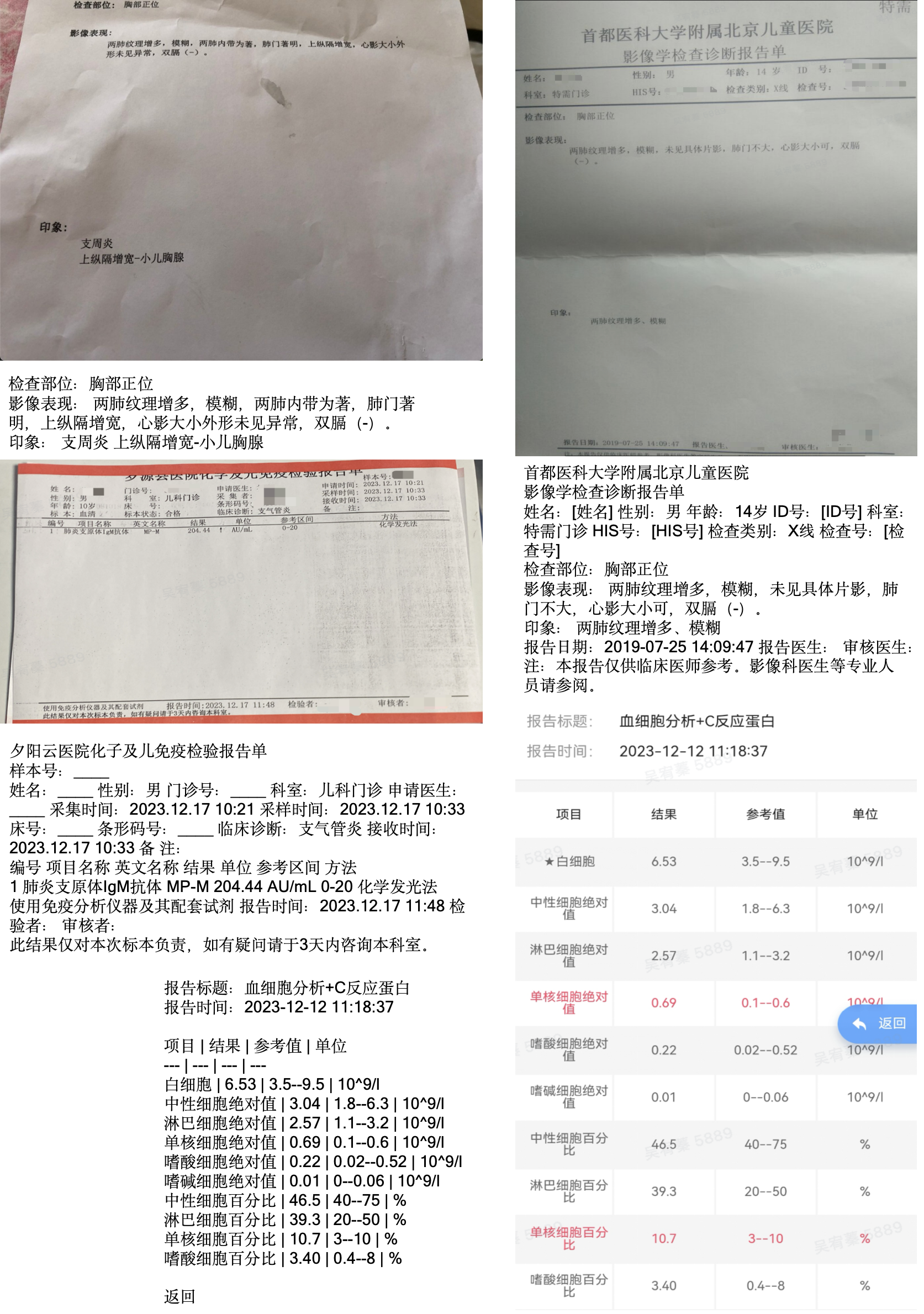}
\caption{Strong OCR ability of Ocean-OCR-3B. Our model shows strong ability for medical report texts recognition. We simply use \textit{Please extract all texts in this image.} as prompt.}
\label{Fig7:med}
\end{figure}

\section{Experiment}
In this section, we carry out a series of experiments to validate the effectiveness of our proposed approaches and demonstrate the strengths in benchmark accuracy and real-world OCR scenarios.

\subsection{General Benchmarks}
To evaluate the general performance of Ocean-OCR, we use comprehensive benchmarks, including MMMU \cite{mmmu}, MMBench-EN \cite{mmbench}, MMBench-CN \cite{mmbench}, 
MathVista \cite{mathvista}, MME \cite{mme}, SEEDBench \cite{seedbench}, RealWorldQA \cite{realworldqa}, and HallusionBench \cite{hallubench}. To ensure consistent and reproducible evaluation results, we uniformly employ VLMEvalKit \cite{duan2024vlmevalkit} for all evaluations. Every evaluation is performed in a zero-shot setting, following the models' original configurations to maintain fairness and uniformity across various models and benchmarks. In Table \ref{table:general_benchmark}, Ocean-OCR shows promising performance compared to other models with a similar number of parameters ($\leq$ 4B). In particular, we find Ocean-OCR has extraordinary performance on SEEDBench and Hallusion Bench.

\begin{table*}[!t]
    \centering
    \renewcommand\arraystretch{1.5}
    \setlength{\tabcolsep}{0.5mm}{
    \footnotesize
    \begin{tabular}{lcccccccc}
        \toprule
        Model   & \makecell[c]{MMMU\\-val} & \makecell[c]{MMBench\\-EN} & \makecell[c]{MMBench\\-CN} & \makecell[c]{MathVista\\-mini} & MME & \makecell[c]{SEED\\Bench} & \makecell[c]{RealWorld\\QA} & \makecell[c]{Hallusion\\Bench}\\
        \midrule
        MM1.5-3B \cite{zhang2024mm1} & 37.1 & - & - & 44.4 & 1798.0 & 72.4 & 56.9 & - \\
        Qwen-2-VL-2B \cite{qwen2} & 40.0 & 72.2 & 70.1 & 43.2 & 1890.0 & 72.8 & 63.1 & 38.8\\
        InternVL-2.5-2B \cite{internvl2d5} & 43.6 & 74.7 & 71.9 &  51.3 & 2138.2 & - & 60.1 & 42.6 \\
        TextHawk2-7B \cite{yu2024texthawk2} & 45.0 & 77.5 & 77.6 & 54.5 & 2125.9 & 74.3 & 66.8 & 49.5 \\
        BlueLM-3B \cite{bluelm} & 45.1 & - & - & 60.8 & - & - & 66.7 & 48 \\
        Megrez-3B-Omni \cite{Megrez-3B} & 51.9 & 80.8 & 82.3 & 62 & 2315 & - & 71.9 & 50.1 \\
        Phi-3.5-Vision-4B* \cite{abdin2024phi} & 44.0 & 74.1 & 59.9 & 44.7 & 1531.6 & 70.9 & 58.7 & 39.8\\
        InternVL-2.5-4B \cite{internvl2d5} & 52.3 & 81.1 & 79.3 & 60.5 & 2337.5 & - & 64.3 & 46.3 \\
        \midrule
        \rowcolor{gray!20}
        Ocean-OCR(Ours) & 42.0 & 75.3 & 73.0 & 55.6 & 2094 & 72.5 & 61.2 & 46.0\\
        \bottomrule
    \end{tabular}
}
\vspace{6pt}
    \caption{\textbf{Comparison of performance on general benchmarks.} * denotes the results are reproduced by ourselves and the others denote officially reported results.}
    \label{table:general_benchmark}
\end{table*}

\subsection{OCR Benchmarks}
To demonstrate the superior OCR ability of our Ocean-OCR, we evalute the performance on representative open-source benchmarks related to OCR, including DocVQA \cite{mathew2021docvqa}, TextVQA \cite{textvqa}, ChartQA \cite{chartqa} and OCRBench \cite{ocrbench}. These open-source OCR benchmarks assess the OCR capabilities across various dimensions. The experimental results in Table \ref{table:ocr_benchmark} demonstrate that our model significantly outperforms other models with comparable parameter sizes in OCR tasks.

\begin{table}[!t]
    \centering
    \renewcommand\arraystretch{1.5}
    \setlength{\tabcolsep}{2mm}{
    \footnotesize
    \begin{tabular}{lcccc>{\columncolor{mylight}}c}
        \toprule
        Model & DocVQA  & TextVQA & ChartQA & OCRBench & Average \\
        \midrule
        MM1.5-3B \cite{zhang2024mm1} & 87.7 & 76.5 & 74.2 & 65.7 & 76.0 \\
        Phi-3.5-Vision-4B* \cite{abdin2024phi}& 84.4 & 73.3 & 81.2 & 63.9 & 75.7 \\
        Qwen-2-VL-2B \cite{qwen2} & 90.1 & 79.6 & 73.4 & 78.3 & 80.4 \\
        InternVL-2.5-2B \cite{internvl2d5}& 88.7 & 74.3 & 79.2 & 80.4 & 80.7\\
        TextHawk2-7B \cite{yu2024texthawk2}& 89.6 & 75.1 & 81.4 & 78.4 & 81.1 \\
        BlueLM-3B \cite{bluelm}& 87.8 & 78.4 & 80.4 & 82.9 & 82.4 \\
        InternVL-2.5-4B \cite{internvl2d5}& 91.6 & 76.8 & 84.0 & 82.8 & 83.8\\
        Megrez-3B-Omni \cite{Megrez-3B}& 91.6 & 80.3 & - & 82.8 & - \\
        \midrule
        \rowcolor{gray!20}
        Ocean-OCR(Ours) & 91.4 & 80.0 & 84.6 & 82.7 & 84.7\\

        \bottomrule
    \end{tabular}
}
\vspace{6pt}
    \caption{\textbf{Comparison of performance on OCR benchmarks.} * denotes the results are reproduced by ourselves and the others denote officially reported results.}
    \label{table:ocr_benchmark}
\end{table}

\subsection{OCR practical scenarios}
In this section, we verify the performance of Ocean-OCR on 4 different OCR practical scenarios, including (1) document understanding; (2) scene text recognition; (3) handwritten recognition.
Note that for each benchmark, the test data is carefully filtered for text similarity to ensure it does not appear in the training data.
GOT \cite{got} is a MLLM designed specifically for OCR tasks. It cannot follow complex instructions and is therefore not compatible with general-purpose tasks. TextIn\footnote{For document extraction we use https://api.textin.com/ai/service/v1/pdf\_to\_markdown, for scene text and handwritten recognition we use https://api.textin.com/ai/service/v2/recognize/multipage} and PaddleOCR\footnote{We download model weights in https://github.com/PaddlePaddle/PaddleOCR for corresponding scenarios} are well-known specialized models in the field of OCR.

\subsubsection{Document extraction}

To assess the OCR capability in understanding bilingual dense document images, we compiled an evaluation dataset consisting of 100 images from English papers and 100 images from Chinese papers. We evaluated the model using a comprehensive set of metrics, including Normalized Edit Distance, F1 Score, Precision, Recall, BLEU, and METEOR. The evaluation results are shown in Table \ref{table:paper_ocr}. Ocean-OCR demonstrates excellent performance in PDF text recognition and document understanding, highlighting its strong capabilities in this OCR task.
Fig. \ref{Fig6:pdf_ocr} shows two cases about document information extraction.
Besides, we find that our Ocean-OCR also has exellent performance in text extraction in medical report image.
We show some cases in Fig. \ref{Fig7:med}.
In our future work, we will establish a benchmark related to the extraction of textual information from medical reports.
This initiative aims to evaluate the performance of mainstream MLLMs as well as specialized OCR models.

\begin{table*}[!t]
    \centering
    \renewcommand\arraystretch{1.5}
    \setlength{\tabcolsep}{1.2mm}{
    \footnotesize
    \begin{tabular}{lcccccccccccc}
        \toprule
        \multirow{2}{*}{Model}   & \multicolumn{2}{c}{Edit Distance $\downarrow$}  & \multicolumn{2}{c}{F1-score $\uparrow$} & \multicolumn{2}{c}{Precision$\uparrow$} & \multicolumn{2}{c}{Recall$\uparrow$} & \multicolumn{2}{c}{BLEU$\uparrow$} & \multicolumn{2}{c}{METEOR$\uparrow$} \\
        \cmidrule(r){2-3} \cmidrule(r){4-5} \cmidrule(r){6-7} \cmidrule(r){8-9} \cmidrule(r){10-11} \cmidrule(r){12-13}  
        & en & zh & en & zh & en & zh & en & zh & en & zh & en & zh\\
        
        \midrule
        InternVL-2.5-2B \cite{internvl2d5}& 0.328 & 0.616 & 0.725 & 0.599 & 0.691 & 0.555 & 0.799 & 0.675 & 0.581 & 0.269 & 0.710 & 0.443 \\
        Phi-3.5-Vision-4B \cite{abdin2024phi}& 0.295 & - &  0.807 & - & 0.785 & - & 0.863 & - & 0.704 & - & 0.816 & - \\
        InternVL-2.5-4B \cite{internvl2d5}& 0.304 & 0.690 & 0.756 & 0.526 & 0.729 & 0.471 & 0.809 & 0.644 & 0.622 & 0.218 & 0.744 & 0.386 \\
        Qwen-2-VL-7B \cite{qwen2}& 0.165 & 0.270 & 0.849 & 0.883 & 0.834 & 0.847 & 0.873 & 0.942 & 0.795 & 0.578 & 0.859 & 0.763 \\
        MiniCPM-V2.6-8B \cite{yao2024minicpm} & 0.244 & 0.437 & 0.804 & 0.778 & 0.793 & 0.721 & 0.837 & 0.875 & 0.695 & 0.431 & 0.640 & 0.642 \\
        \midrule
        GOT \cite{got} & 0.084 & 0.117 & 0.895 & 0.928 & 0.891 & 0.934 & 0.906 & 0.929 &0.835 & 0.805 & 0.874 & 0.848\\
        TextIn & 0.055 & 0.217 & 0.861 & 0.936 & 0.856 & 0.948 & 0.866 & 0.924 &0.773 & 0.566 & 0.887 & 0.782\\
        PaddleOCR & 0.323 & 0.649 & 0.707 & 0.864 & 0.690 & 0.912 & 0.730 & 0.821 & 0.517 & 0.537 & 0.674 & 0.699 \\
        \midrule
        \rowcolor{gray!20}
        Ocean-OCR(Ours) & 0.057 & 0.062 & 0.937 & 0.962 & 0.932 & 0.956 & 0.956 & 0.974 & 0.906 & 0.912 & 0.945 &0.916\\

        \bottomrule
    \end{tabular}
}    
\vspace{6pt}
\caption{\textbf{Comparison of performance on dense English(en) and Chinese(zh) OCR for document-level pages.} Phi-3.5-Vision-4B fails in following instruction on Chinese document-level pages.}
    \label{table:paper_ocr}
\end{table*}

\subsubsection{Scene text recognition}
Scene text is ubiquitous in daily life, found on everything from street signs to product packaging, highlighting the crucial need for accurate recognition of scene text. Effective scene text recognition not only enhances the accessibility of information but also plays a vital role in various applications, from assistive technologies to automated systems.
We have assembled a scene text OCR benchmark comprising 260 natural images, evenly divided between Chinese and English.
These images are sampled from MSRA-TD500-Dataset \cite{yao2012detecting}.
Each image is first passed through PaddleOCR tools to get pseudo ground truth, and then each one has been manually verified and corrected to ensure the accuracy of the ground truth annotations. As illustrated in Table \ref{table:scene_ocr}, Ocean-OCR performs outstandingly in natural scene OCR tasks, accurately identifying text even when it comprises only a small part of the image.
For example, Ocean-OCR with 3B parameters even surpasses Qwen2-VL-7B and MiniCPM-V2.6-8B in all the six indicators, such as \textit{Edit Distance} (0.163 in Qwen2-VL-7B, 0.146 in MiniCPM-V2.6-8B, and 0.113 in Ocean-OCR) and \textit{METEOR} (0.752 in Qwen2-VL-7B, 0.734 in MiniCPM-V2.6-8B, and 0.754 in Ocean-OCR).
This underscores the model's capability to handle intricate and varied real-world scenes with high accuracy.
We show some cases in Fig. \ref{Fig5:scene_ocr}.

\begin{table*}[!t]
    \centering
    \renewcommand\arraystretch{1.5}
    \setlength{\tabcolsep}{1mm}{
    \footnotesize
    \begin{tabular}{lcccccc}
        \toprule
        Model & Edit Distance$\downarrow$ & F1-score $\uparrow$ & Precision $\uparrow$ & Recall$\uparrow$ & BLEU$\uparrow$ & METEOR$\uparrow$ \\
        \midrule
        Qwen-2-VL-2B \cite{qwen2}& 0.292 & 0.710 & 0.705 & 0.757 & 0.283 & 0.641 \\
        InternVL-2.5-2B \cite{internvl2d5}& 0.193 & 0.807 & 0.807 & 0.824 & 0.293 & 0.683 \\
        Phi-3.5-Vision-4B \cite{abdin2024phi}& 0.452 & 0.595 & 0.498 & 0.595 & 0.152 & 0.398 \\
        InternVL-2.5-4B \cite{internvl2d5}& 0.184 & 0.820 & 0.834 & 0.820 & 0.328 & 0.683 \\
        Qwen-2-VL-7B \cite{qwen2}& 0.163 & 0.835 & 0.827 & 0.865 & 0.380 & 0.752 \\
        MiniCPM-V2.6-8B \cite{yao2024minicpm}& 0.146 & 0.857 & 0.844 & 0.887 & 0.372 & 0.734\\
        \midrule
        GOT \cite{got}& 0.251 & 0.702 & 0.689 & 0.748 & 0.241 & 0.610 \\
        TextIn & 0.213 & 0.725 & 0.752 & 0.712 & 0.352 & 0.629 \\
        PaddleOCR & 0.130 & 0.837 & 0.828 & 0.858 & 0.387 & 0.720 \\
        \midrule
        \rowcolor{gray!20}
        Ocean-OCR(Ours) & 0.113 & 0.875 & 0.875 & 0.887 & 0.420 & 0.754 \\
        \bottomrule
    \end{tabular}
}
\vspace{6pt}
    \caption{\textbf{Comparison of performance on OCR for scene texts.}}
    \label{table:scene_ocr}
\end{table*}

\subsubsection{Handwritten recognition}

Handwritten text recognition is also a crucial component in evaluating a model's OCR capabilities. To provide a thorough assessment, we have developed a multi-granularity handwritten text recognition evaluation dataset that incorporates both real and synthetic bilingual data. This dataset specifically includes: (1) Paragraph-level real Chinese and English data (from CASIA-HWDB \cite{liu2011casia} and GNHK \cite{lee2021gnhk}); (2) Line-level real Chinese and English data (from CASIA-HWDB and BRUSH \cite{kotani2020generating}); (3) Paragraph-level synthetic Chinese and English data; (4) Line-level synthetic Chinese and English data. Each category contains 100 samples. On the constructed dataset, the evaluation metrics we use include Normalized Edit Distance, F1 Score, Precision, Recall, BLEU, and METEOR. As demonstrated in Table \ref{table:handwritten_ocr}, Ocean-OCR also shows impressive performance in challenging tasks such as handwritten text recognition, demonstrating its robust capabilities in handling complex and varied handwriting styles.
In Fig. \ref{Fig6:handwritten_ocr}, we show several cases of handwritten recognition in Chinese and English.

\begin{table*}[!t]
    \centering
    \renewcommand\arraystretch{1.5}
    \setlength{\tabcolsep}{1.2mm}{
    \footnotesize
    \begin{tabular}{lcccccccccccc}
        \toprule
        \multirow{2}{*}{Model}   & \multicolumn{2}{c}{Edit Distance $\downarrow$}  & \multicolumn{2}{c}{F1-score $\uparrow$} & \multicolumn{2}{c}{Precision$\uparrow$} & \multicolumn{2}{c}{Recall$\uparrow$} & \multicolumn{2}{c}{BLEU$\uparrow$} & \multicolumn{2}{c}{METEOR$\uparrow$} \\
        \cmidrule(r){2-3} \cmidrule(r){4-5} \cmidrule(r){6-7} \cmidrule(r){8-9} \cmidrule(r){10-11} \cmidrule(r){12-13}  
        & en & zh & en & zh & en & zh & en & zh & en & zh & en & zh\\
        
        \midrule
        InternVL-2.5-2B \cite{internvl2d5}& 0.227 & 0.255 & 0.619 & 0.717 & 0.633 & 0.733 & 0.614 & 0.709 & 0.363 & 0.464 & 0.611 & 0.661\\
        InternVL-2.5-4B \cite{internvl2d5}& 0.197 & 0.240 & 0.661 & 0.741 & 0.674 & 0.754 & 0.655 & 0.734 & 0.406 & 0.473 & 0.652 & 0.687 \\
        Qwen-2-VL-7B \cite{qwen2}& 0.127 & 0.113 & 0.760 & 0.881 & 0.773 & 0.884 &  0.754 & 0.884 &  0.490 & 0.666 & 0.756 & 0.859\\
        MiniCPM-V2.6-8B \cite{yao2024minicpm}& 0.147 & 0.175 & 0.727 & 0.810 & 0.747 & 0.811 & 0.714 & 0.812 & 0.443 & 0.583 & 0.727 & 0.774\\
        \midrule
        GOT \cite{got}& 0.616 & 0.402 & 0.283 & 0.568 & 0.309 & 0.618 & 0.273 & 0.544 & 0.151 & 0.295 & 0.255 & 0.492 \\
        TextIn & 0.358 & 0.180 & 0.362 & 0.840 & 0.368 & 0.869 & 0.362 & 0.822 & 0.098 & 0.567 & 0.337 & 0.751 \\
        PaddleOCR & 0.418 & 0.325 & 0.237 & 0.664 & 0.232 & 0.646 & 0.263 & 0.700 & 0.069 & 0.431 & 0.236 & 0.648 \\ 
        \midrule
        \rowcolor{gray!20}
        Ocean-OCR(Ours) & 0.145 & 0.106 & 0.774 & 0.885 & 0.780 & 0.912 & 0.782 & 0.862 & 0.532 & 0.736 & 0.772 & 0.885 \\

        \bottomrule
    \end{tabular}
}
    \caption{\textbf{Comparison of performance on English(en) and Chinese(zh) OCR for handwritten recognition.} We find that Qwen-2-VL-2B and Phi-3.5-Vision-4B have trouble following instructions in this scenario.}
    \label{table:handwritten_ocr}
\end{table*}

\section{Conclusion}
In this study, we introduced Ocean-OCR, a 3B MLLM that addresses the OCR limitations of existing multimodal models. Using NaViT, Ocean-OCR handles variable resolution inputs effectively, enhancing its adaptability to different image qualities. Trained on high-quality OCR datasets, Ocean-OCR excels in diverse OCR scenarios.
Our experiments on open-source OCR benchmarks and real-world applications demonstrate Ocean-OCR's superior performance and robustness. This work sets a new benchmark for multimodal learning-based OCR tasks, providing a powerful tool for accurate visual-textual information processing.
\bibliographystyle{plain}
\bibliography{ref}

\begin{thebibliography}{10}

\bibitem{RenderedText}
Renderedtext.
\newblock \url{https://huggingface.co/datasets/wendlerc/RenderedText/}, 2024.

\bibitem{abdin2024phi}
Marah Abdin, Jyoti Aneja, Hany Awadalla, Ahmed Awadallah, Ammar~Ahmad Awan, Nguyen Bach, Amit Bahree, Arash Bakhtiari, Jianmin Bao, Harkirat Behl, et~al.
\newblock Phi-3 technical report: A highly capable language model locally on your phone.
\newblock {\em arXiv preprint arXiv:2404.14219}, 2024.

\bibitem{Megrez-3B}
Infinigence AI.
\newblock Megrez-3b-omni.
\newblock \url{https://huggingface.co/Infinigence/Megrez-3B-Omni/}, 2024.

\bibitem{flamingo}
Jean-Baptiste Alayrac, Jeff Donahue, Pauline Luc, Antoine Miech, Iain Barr, Yana Hasson, Karel Lenc, Arthur Mensch, Katherine Millican, Malcolm Reynolds, et~al.
\newblock Flamingo: a visual language model for few-shot learning.
\newblock {\em Advances in neural information processing systems}, 35:23716--23736, 2022.

\bibitem{internvl2d5}
Zhe Chen, Weiyun Wang, Yue Cao, Yangzhou Liu, Zhangwei Gao, Erfei Cui, Jinguo Zhu, Shenglong Ye, Hao Tian, Zhaoyang Liu, et~al.
\newblock Expanding performance boundaries of open-source multimodal models with model, data, and test-time scaling.
\newblock {\em arXiv preprint arXiv:2412.05271}, 2024.

\bibitem{internvl1d5}
Zhe Chen, Weiyun Wang, Hao Tian, Shenglong Ye, Zhangwei Gao, Erfei Cui, Wenwen Tong, Kongzhi Hu, Jiapeng Luo, Zheng Ma, et~al.
\newblock How far are we to gpt-4v? closing the gap to commercial multimodal models with open-source suites.
\newblock {\em Science China Information Sciences}, 67(12):220101, 2024.

\bibitem{chen2024far}
Zhe Chen, Weiyun Wang, Hao Tian, Shenglong Ye, Zhangwei Gao, Erfei Cui, Wenwen Tong, Kongzhi Hu, Jiapeng Luo, Zheng Ma, et~al.
\newblock How far are we to gpt-4v? closing the gap to commercial multimodal models with open-source suites.
\newblock {\em arXiv preprint arXiv:2404.16821}, 2024.

\bibitem{internvl}
Zhe Chen, Jiannan Wu, Wenhai Wang, Weijie Su, Guo Chen, Sen Xing, Muyan Zhong, Qinglong Zhang, Xizhou Zhu, Lewei Lu, et~al.
\newblock Internvl: Scaling up vision foundation models and aligning for generic visual-linguistic tasks.
\newblock In {\em Proceedings of the IEEE/CVF Conference on Computer Vision and Pattern Recognition}, pages 24185--24198, 2024.

\bibitem{chen2023internvl}
Zhe Chen, Jiannan Wu, Wenhai Wang, Weijie Su, Guo Chen, Sen Xing, Muyan Zhong, Qinglong Zhang, Xizhou Zhu, Lewei Lu, Bin Li, Ping Luo, Tong Lu, Yu~Qiao, and Jifeng Dai.
\newblock Internvl: Scaling up vision foundation models and aligning for generic visual-linguistic tasks.
\newblock {\em arXiv preprint arXiv:2312.14238}, 2023.

\bibitem{chng2019icdar2019}
Chee~Kheng Chng, Yuliang Liu, Yipeng Sun, Chun~Chet Ng, Canjie Luo, Zihan Ni, ChuanMing Fang, Shuaitao Zhang, Junyu Han, Errui Ding, et~al.
\newblock Icdar2019 robust reading challenge on arbitrary-shaped text-rrc-art.
\newblock In {\em 2019 International Conference on Document Analysis and Recognition (ICDAR)}, pages 1571--1576. IEEE, 2019.

\bibitem{realworldqa}
X.AI Corp.
\newblock Grok-1.5 vision preview: Connecting the digital and physical worlds with our first multimodal model., 2024.

\bibitem{navit}
Mostafa Dehghani, Basil Mustafa, Josip Djolonga, Jonathan Heek, Matthias Minderer, Mathilde Caron, Andreas Steiner, Joan Puigcerver, Robert Geirhos, Ibrahim~M Alabdulmohsin, et~al.
\newblock Patch n’pack: Navit, a vision transformer for any aspect ratio and resolution.
\newblock {\em Advances in Neural Information Processing Systems}, 36, 2024.

\bibitem{dong2024baichuanseed}
Guosheng Dong, Da~Pan, Yiding Sun, Shusen Zhang, Zheng Liang, Xin Wu, Yanjun Shen, Fan Yang, Haoze Sun, Tianpeng Li, et~al.
\newblock Baichuanseed: Sharing the potential of extensive data collection and deduplication by introducing a competitive large language model baseline.
\newblock {\em arXiv preprint arXiv:2408.15079}, 2024.

\bibitem{internlm}
Xiaoyi Dong, Pan Zhang, Yuhang Zang, Yuhang Cao, Bin Wang, Linke Ouyang, Songyang Zhang, Haodong Duan, Wenwei Zhang, Yining Li, et~al.
\newblock Internlm-xcomposer2-4khd: A pioneering large vision-language model handling resolutions from 336 pixels to 4k hd.
\newblock {\em arXiv preprint arXiv:2404.06512}, 2024.

\bibitem{du2021pp}
Yuning Du, Chenxia Li, Ruoyu Guo, Cheng Cui, Weiwei Liu, Jun Zhou, Bin Lu, Yehua Yang, Qiwen Liu, Xiaoguang Hu, et~al.
\newblock Pp-ocrv2: Bag of tricks for ultra lightweight ocr system.
\newblock {\em arXiv preprint arXiv:2109.03144}, 2021.

\bibitem{duan2024vlmevalkit}
Haodong Duan, Junming Yang, Yuxuan Qiao, Xinyu Fang, Lin Chen, Yuan Liu, Xiaoyi Dong, Yuhang Zang, Pan Zhang, Jiaqi Wang, et~al.
\newblock Vlmevalkit: An open-source toolkit for evaluating large multi-modality models.
\newblock In {\em Proceedings of the 32nd ACM International Conference on Multimedia}, pages 11198--11201, 2024.

\bibitem{mme}
Chaoyou Fu, Peixian Chen, Yunhang Shen, Yulei Qin, Mengdan Zhang, Xu~Lin, Zhenyu Qiu, Wei Lin, Jinrui Yang, Xiawu Zheng, Li~Ke, Sun Xing, Wu~Yunsheng, and Ji~Rongrong.
\newblock Mme: A comprehensive evaluation benchmark for multimodal large language models.
\newblock {\em arXiv preprint arXiv:2306.13394}, 2023.

\bibitem{gu2022wukong}
Jiaxi Gu, Xiaojun Meng, Guansong Lu, Lu~Hou, Niu Minzhe, Xiaodan Liang, Lewei Yao, Runhui Huang, Wei Zhang, Xin Jiang, et~al.
\newblock Wukong: A 100 million large-scale chinese cross-modal pre-training benchmark.
\newblock {\em Advances in Neural Information Processing Systems}, 35:26418--26431, 2022.

\bibitem{hallubench}
Tianrui Guan, Fuxiao Liu, Xiyang Wu, Ruiqi Xian, Zongxia Li, Xiaoyu Liu, Xijun Wang, Lichang Chen, Furong Huang, Yaser Yacoob, et~al.
\newblock Hallusionbench: an advanced diagnostic suite for entangled language hallucination and visual illusion in large vision-language models.
\newblock In {\em Proceedings of the IEEE/CVF Conference on Computer Vision and Pattern Recognition}, pages 14375--14385, 2024.

\bibitem{mplug1d5}
Anwen Hu, Haiyang Xu, Jiabo Ye, Ming Yan, Liang Zhang, Bo~Zhang, Chen Li, Ji~Zhang, Qin Jin, Fei Huang, et~al.
\newblock mplug-docowl 1.5: Unified structure learning for ocr-free document understanding.
\newblock {\em arXiv preprint arXiv:2403.12895}, 2024.

\bibitem{hu2024mplug}
Anwen Hu, Haiyang Xu, Jiabo Ye, Ming Yan, Liang Zhang, Bo~Zhang, Chen Li, Ji~Zhang, Qin Jin, Fei Huang, et~al.
\newblock mplug-docowl 1.5: Unified structure learning for ocr-free document understanding.
\newblock {\em arXiv preprint arXiv:2403.12895}, 2024.

\bibitem{mplug2}
Anwen Hu, Haiyang Xu, Liang Zhang, Jiabo Ye, Ming Yan, Ji~Zhang, Qin Jin, Fei Huang, and Jingren Zhou.
\newblock mplug-docowl2: High-resolution compressing for ocr-free multi-page document understanding.
\newblock {\em arXiv preprint arXiv:2409.03420}, 2024.

\bibitem{kim2022donut}
Geewook Kim, Teakgyu Hong, Moonbin Yim, JeongYeon Nam, Jinyoung Park, Jinyeong Yim, Wonseok Hwang, Sangdoo Yun, Dongyoon Han, and Seunghyun Park.
\newblock Ocr-free document understanding transformer.
\newblock In {\em European Conference on Computer Vision (ECCV)}, 2022.

\bibitem{sam}
Alexander Kirillov, Eric Mintun, Nikhila Ravi, Hanzi Mao, Chloe Rolland, Laura Gustafson, Tete Xiao, Spencer Whitehead, Alexander~C Berg, Wan-Yen Lo, et~al.
\newblock Segment anything.
\newblock In {\em Proceedings of the IEEE/CVF International Conference on Computer Vision}, pages 4015--4026, 2023.

\bibitem{kotani2020generating}
Atsunobu Kotani, Stefanie Tellex, and James Tompkin.
\newblock Generating handwriting via decoupled style descriptors.
\newblock In {\em European Conference on Computer Vision}, pages 764--780. Springer, 2020.

\bibitem{laurenccon2024obelics}
Hugo Lauren{\c{c}}on, Lucile Saulnier, L{\'e}o Tronchon, Stas Bekman, Amanpreet Singh, Anton Lozhkov, Thomas Wang, Siddharth Karamcheti, Alexander Rush, Douwe Kiela, et~al.
\newblock Obelics: An open web-scale filtered dataset of interleaved image-text documents.
\newblock {\em Advances in Neural Information Processing Systems}, 36, 2024.

\bibitem{laurençon2024matters}
Hugo Laurençon, Léo Tronchon, Matthieu Cord, and Victor Sanh.
\newblock What matters when building vision-language models?, 2024.

\bibitem{lee2021gnhk}
Alex~WC Lee, Jonathan Chung, and Marco Lee.
\newblock Gnhk: a dataset for english handwriting in the wild.
\newblock In {\em Document Analysis and Recognition--ICDAR 2021: 16th International Conference, Lausanne, Switzerland, September 5--10, 2021, Proceedings, Part IV 16}, pages 399--412. Springer, 2021.

\bibitem{seedbench}
Bohao Li, Yuying Ge, Yixiao Ge, Guangzhi Wang, Rui Wang, Ruimao Zhang, and Ying Shan.
\newblock Seed-bench: Benchmarking multimodal large language models.
\newblock In {\em Proceedings of the IEEE/CVF Conference on Computer Vision and Pattern Recognition}, pages 13299--13308, 2024.

\bibitem{blip2}
Junnan Li, Dongxu Li, Silvio Savarese, and Steven Hoi.
\newblock Blip-2: Bootstrapping language-image pre-training with frozen image encoders and large language models.
\newblock In {\em International conference on machine learning}, pages 19730--19742. PMLR, 2023.

\bibitem{li2024densefusion}
Xiaotong Li, Fan Zhang, Haiwen Diao, Yueze Wang, Xinlong Wang, and Ling-Yu Duan.
\newblock Densefusion-1m: Merging vision experts for comprehensive multimodal perception.
\newblock {\em arXiv preprint arXiv:2407.08303}, 2024.

\bibitem{minigemini}
Yanwei Li, Yuechen Zhang, Chengyao Wang, Zhisheng Zhong, Yixin Chen, Ruihang Chu, Shaoteng Liu, and Jiaya Jia.
\newblock Mini-gemini: Mining the potential of multi-modality vision language models.
\newblock {\em arXiv preprint arXiv:2403.18814}, 2024.

\bibitem{monkey}
Zhang Li, Biao Yang, Qiang Liu, Zhiyin Ma, Shuo Zhang, Jingxu Yang, Yabo Sun, Yuliang Liu, and Xiang Bai.
\newblock Monkey: Image resolution and text label are important things for large multi-modal models.
\newblock {\em CoRR}, abs/2311.06607, 2023.

\bibitem{vila}
Ji~Lin, Hongxu Yin, Wei Ping, Pavlo Molchanov, Mohammad Shoeybi, and Song Han.
\newblock Vila: On pre-training for visual language models.
\newblock In {\em Proceedings of the IEEE/CVF Conference on Computer Vision and Pattern Recognition}, pages 26689--26699, 2024.

\bibitem{liu2011casia}
Cheng-Lin Liu, Fei Yin, Da-Han Wang, and Qiu-Feng Wang.
\newblock Casia online and offline chinese handwriting databases.
\newblock In {\em 2011 international conference on document analysis and recognition}, pages 37--41. IEEE, 2011.

\bibitem{llava1d5}
Haotian Liu, Chunyuan Li, Yuheng Li, and Yong~Jae Lee.
\newblock Improved baselines with visual instruction tuning.
\newblock In {\em Proceedings of the IEEE/CVF Conference on Computer Vision and Pattern Recognition}, pages 26296--26306, 2024.

\bibitem{llava_next}
Haotian Liu, Chunyuan Li, Yuheng Li, Bo~Li, Yuanhan Zhang, Sheng Shen, and Yong~Jae Lee.
\newblock Llava-next: Improved reasoning, ocr, and world knowledge, 2024.

\bibitem{llava}
Haotian Liu, Chunyuan Li, Qingyang Wu, and Yong~Jae Lee.
\newblock Visual instruction tuning.
\newblock {\em Advances in neural information processing systems}, 36, 2024.

\bibitem{mmbench}
Yuan Liu, Haodong Duan, Yuanhan Zhang, Bo~Li, Songyang Zhang, Wangbo Zhao, Yike Yuan, Jiaqi Wang, Conghui He, Ziwei Liu, et~al.
\newblock Mmbench: Is your multi-modal model an all-around player?
\newblock In {\em European conference on computer vision}, pages 216--233. Springer, 2025.

\bibitem{liu2024points1}
Yuan Liu, Le~Tian, Xiao Zhou, Xinyu Gao, Kavio Yu, Yang Yu, and Jie Zhou.
\newblock Points1. 5: Building a vision-language model towards real world applications.
\newblock {\em arXiv preprint arXiv:2412.08443}, 2024.

\bibitem{ocrbench}
Yuliang Liu, Zhang Li, Biao Yang, Chunyuan Li, Xucheng Yin, Cheng-lin Liu, Lianwen Jin, and Xiang Bai.
\newblock On the hidden mystery of ocr in large multimodal models.
\newblock {\em arXiv preprint arXiv:2305.07895}, 2023.

\bibitem{textmonkey}
Yuliang Liu, Biao Yang, Qiang Liu, Zhang Li, Zhiyin Ma, Shuo Zhang, and Xiang Bai.
\newblock Textmonkey: An ocr-free large multimodal model for understanding document.
\newblock {\em arXiv preprint arXiv:2403.04473}, 2024.

\bibitem{lu2024datasculpt}
Keer Lu, Zheng Liang, Xiaonan Nie, Da~Pan, Shusen Zhang, Keshi Zhao, Weipeng Chen, Zenan Zhou, Guosheng Dong, Wentao Zhang, et~al.
\newblock Datasculpt: Crafting data landscapes for llm post-training through multi-objective partitioning.
\newblock {\em arXiv preprint arXiv:2409.00997}, 2024.

\bibitem{mathvista}
Pan Lu, Hritik Bansal, Tony Xia, Jiacheng Liu, Chunyuan Li, Hannaneh Hajishirzi, Hao Cheng, Kai-Wei Chang, Michel Galley, and Jianfeng Gao.
\newblock Mathvista: Evaluating mathematical reasoning of foundation models in visual contexts.
\newblock {\em arXiv preprint arXiv:2310.02255}, 2023.

\bibitem{bluelm}
Xudong Lu, Yinghao Chen, Cheng Chen, Hui Tan, Boheng Chen, Yina Xie, Rui Hu, Guanxin Tan, Renshou Wu, Yan Hu, et~al.
\newblock Bluelm-v-3b: Algorithm and system co-design for multimodal large language models on mobile devices.
\newblock {\em arXiv preprint arXiv:2411.10640}, 2024.

\bibitem{chartqa}
Ahmed Masry, Do~Xuan Long, Jia~Qing Tan, Shafiq Joty, and Enamul Hoque.
\newblock Chartqa: A benchmark for question answering about charts with visual and logical reasoning.
\newblock {\em arXiv preprint arXiv:2203.10244}, 2022.

\bibitem{mathew2021docvqa}
Minesh Mathew, Dimosthenis Karatzas, and CV~Jawahar.
\newblock Docvqa: A dataset for vqa on document images.
\newblock In {\em Proceedings of the IEEE/CVF winter conference on applications of computer vision}, pages 2200--2209, 2021.

\bibitem{gpt4o}
OpenAI.
\newblock Hello gpt-4o, 2024.

\bibitem{clip}
Alec Radford, Jong~Wook Kim, Chris Hallacy, Aditya Ramesh, Gabriel Goh, Sandhini Agarwal, Girish Sastry, Amanda Askell, Pamela Mishkin, Jack Clark, et~al.
\newblock Learning transferable visual models from natural language supervision.
\newblock In {\em International conference on machine learning}, pages 8748--8763. PMLR, 2021.

\bibitem{schuhmann2022laion}
Christoph Schuhmann, Romain Beaumont, Richard Vencu, Cade Gordon, Ross Wightman, Mehdi Cherti, Theo Coombes, Aarush Katta, Clayton Mullis, Mitchell Wortsman, et~al.
\newblock Laion-5b: An open large-scale dataset for training next generation image-text models.
\newblock {\em Advances in Neural Information Processing Systems}, 35:25278--25294, 2022.

\bibitem{textvqa}
Amanpreet Singh, Vivek Natarajan, Meet Shah, Yu~Jiang, Xinlei Chen, Dhruv Batra, Devi Parikh, and Marcus Rohrbach.
\newblock Towards vqa models that can read.
\newblock In {\em Proceedings of the IEEE/CVF conference on computer vision and pattern recognition}, pages 8317--8326, 2019.

\bibitem{qwen2.5}
Qwen Team.
\newblock Qwen2.5: A party of foundation models, September 2024.

\bibitem{tuo2023anytext}
Yuxiang Tuo, Wangmeng Xiang, Jun-Yan He, Yifeng Geng, and Xuansong Xie.
\newblock Anytext: Multilingual visual text generation and editing.
\newblock 2023.

\bibitem{veit2016coco}
Andreas Veit, Tomas Matera, Lukas Neumann, Jiri Matas, and Serge Belongie.
\newblock Coco-text: Dataset and benchmark for text detection and recognition in natural images.
\newblock {\em arXiv preprint arXiv:1601.07140}, 2016.

\bibitem{qwen2}
Peng Wang, Shuai Bai, Sinan Tan, Shijie Wang, Zhihao Fan, Jinze Bai, Keqin Chen, Xuejing Liu, Jialin Wang, Wenbin Ge, et~al.
\newblock Qwen2-vl: Enhancing vision-language model's perception of the world at any resolution.
\newblock {\em arXiv preprint arXiv:2409.12191}, 2024.

\bibitem{qwen2vl}
Peng Wang, Shuai Bai, Sinan Tan, Shijie Wang, Zhihao Fan, Jinze Bai, Keqin Chen, Xuejing Liu, Jialin Wang, Wenbin Ge, et~al.
\newblock Qwen2-vl: Enhancing vision-language model's perception of the world at any resolution.
\newblock {\em arXiv preprint arXiv:2409.12191}, 2024.

\bibitem{vary}
Haoran Wei, Lingyu Kong, Jinyue Chen, Liang Zhao, Zheng Ge, Jinrong Yang, Jianjian Sun, Chunrui Han, and Xiangyu Zhang.
\newblock Vary: Scaling up the vision vocabulary for large vision-language model.
\newblock In {\em European Conference on Computer Vision}, pages 408--424. Springer, 2025.

\bibitem{got}
Haoran Wei, Chenglong Liu, Jinyue Chen, Jia Wang, Lingyu Kong, Yanming Xu, Zheng Ge, Liang Zhao, Jianjian Sun, Yuang Peng, et~al.
\newblock General ocr theory: Towards ocr-2.0 via a unified end-to-end model.
\newblock 2024.

\bibitem{yao2012detecting}
Cong Yao, Xiang Bai, Wenyu Liu, Yi~Ma, and Zhuowen Tu.
\newblock Detecting texts of arbitrary orientations in natural images.
\newblock In {\em 2012 IEEE conference on computer vision and pattern recognition}, pages 1083--1090. IEEE, 2012.

\bibitem{yao2015incidental}
Cong Yao, Jianan Wu, Xinyu Zhou, Chi Zhang, Shuchang Zhou, Zhimin Cao, and Qi~Yin.
\newblock Incidental scene text understanding: Recent progresses on icdar 2015 robust reading competition challenge 4.
\newblock {\em arXiv preprint arXiv:1511.09207}, 2015.

\bibitem{yao2024minicpm}
Yuan Yao, Tianyu Yu, Ao~Zhang, Chongyi Wang, Junbo Cui, Hongji Zhu, Tianchi Cai, Haoyu Li, Weilin Zhao, Zhihui He, et~al.
\newblock Minicpm-v: A gpt-4v level mllm on your phone.
\newblock {\em arXiv preprint arXiv:2408.01800}, 2024.

\bibitem{mplug}
Jiabo Ye, Anwen Hu, Haiyang Xu, Qinghao Ye, Ming Yan, Yuhao Dan, Chenlin Zhao, Guohai Xu, Chenliang Li, Junfeng Tian, et~al.
\newblock mplug-docowl: Modularized multimodal large language model for document understanding.
\newblock {\em arXiv preprint arXiv:2307.02499}, 2023.

\bibitem{survey}
Shukang Yin, Chaoyou Fu, Sirui Zhao, Ke~Li, Xing Sun, Tong Xu, and Enhong Chen.
\newblock A survey on multimodal large language models.
\newblock {\em arXiv preprint arXiv:2306.13549}, 2023.

\bibitem{yu2024texthawk2}
Ya-Qi Yu, Minghui Liao, Jiwen Zhang, and Jihao Wu.
\newblock Texthawk2: A large vision-language model excels in bilingual ocr and grounding with 16x fewer tokens.
\newblock {\em arXiv preprint arXiv:2410.05261}, 2024.

\bibitem{mmmu}
Xiang Yue, Yuansheng Ni, Kai Zhang, Tianyu Zheng, Ruoqi Liu, Ge~Zhang, Samuel Stevens, Dongfu Jiang, Weiming Ren, Yuxuan Sun, et~al.
\newblock Mmmu: A massive multi-discipline multimodal understanding and reasoning benchmark for expert agi.
\newblock In {\em Proceedings of the IEEE/CVF Conference on Computer Vision and Pattern Recognition}, pages 9556--9567, 2024.

\bibitem{zhang2024mm1}
Haotian Zhang, Mingfei Gao, Zhe Gan, Philipp Dufter, Nina Wenzel, Forrest Huang, Dhruti Shah, Xianzhi Du, Bowen Zhang, Yanghao Li, et~al.
\newblock Mm1. 5: Methods, analysis \& insights from multimodal llm fine-tuning.
\newblock {\em arXiv preprint arXiv:2409.20566}, 2024.

\bibitem{DreamLIP}
Kecheng Zheng, Yifei Zhang, Wei Wu, Fan Lu, Shuailei Ma, Xin Jin, Wei Chen, and Yujun Shen.
\newblock Dreamlip: Language-image pre-training with long captions.
\newblock In {\em ECCV}, 2024.

\end{thebibliography}
\end{document}